\documentclass[journal]{IEEEtran}
\usepackage{cite}
\usepackage{amsmath,amssymb,amsfonts}
\usepackage{algorithmic}
\usepackage{graphicx}
\usepackage{textcomp}
\usepackage{csquotes}
\usepackage{xcolor,soul}
\usepackage{adjustbox}
\usepackage{hyperref}
\usepackage{multirow}
\usepackage{titlesec}
\usepackage{subcaption}
\usepackage{makecell}
\usepackage{orcidlink}
\usepackage{booktabs} % for better table rules
\usepackage{caption} % for customizing captions
\usepackage[numbers]{natbib}
\usepackage[ruled,vlined,linesnumbered,boxed]{algorithm2e}
\usepackage{cancel}
\DeclareMathOperator{\argmin}{argmin}
\usepackage{etoolbox}
\usepackage{adjustbox}

\definecolor{columbiablue}{rgb}{0.61, 0.87, 1.0}
\newcommand{\hlcc}[1]{\sethlcolor{white}\hl{#1}}%
\titlespacing{\subsubsection}{0pt}{0pt}{1pt}
\begin{document}

\title{Seamless Underwater Navigation with Limited Doppler Velocity Log Measurements}

\author{Nadav~Cohen \orcidlink{0000-0002-8249-0239}
        and~Itzik~Klein \orcidlink{0000-0001-7846-0654}
% <-this % stops a space
\thanks{N. Cohen and I. Klein are with the Hatter Department of Marine Technologies, Charney School of Marine Sciences, University of Haifa, Israel.\\ Corresponding author: ncohe140@campus.haifa.ac.il (N.Cohen)}% <-this % stops a space
}


\maketitle

\begin{abstract}
\hlcc{Autonomous Underwater Vehicles (AUVs) commonly utilize an inertial navigation system (INS) and a Doppler velocity log (DVL) for underwater navigation. To that end, their measurements are integrated through a nonlinear filter such as the extended Kalman filter (EKF).} The DVL velocity vector estimate depends on retrieving reflections from the seabed, ensuring that at least three out of its four transmitted acoustic beams return successfully. When fewer than three beams are obtained, the DVL cannot provide a velocity update to bind the navigation solution drift. To cope with this challenge, in this paper, we propose a hybrid neural coupled (HNC) approach for seamless AUV navigation in situations of limited DVL measurements. First, we drive an approach to regress two or three missing DVL beams. Then, those beams, together with the measured beams, are incorporated into the EKF. We examined  INS/DVL fusion both in loosely and tightly coupled approaches. Our method was trained and evaluated on recorded data from AUV experiments conducted in the Mediterranean Sea on two different occasions.\hlcc{The results illustrate that our proposed method outperforms the baseline loosely and tightly coupled model-based approaches by an average of 96.15\%. It also demonstrates superior performance compared to a model-based beam estimator by an average of 12.41\% in terms of velocity accuracy for scenarios involving two or three missing beams. Therefore, we demonstrate that our approach offers seamless AUV navigation in situations of limited beam measurements.}
\end{abstract}

% Note that keywords are not normally used for peer-reviewed papers.

\IEEEpeerreviewmaketitle

\section{Introduction}
\begin{figure*}[h!] % [t] option to align at the top
    \centering
    \includegraphics[width=\textwidth]{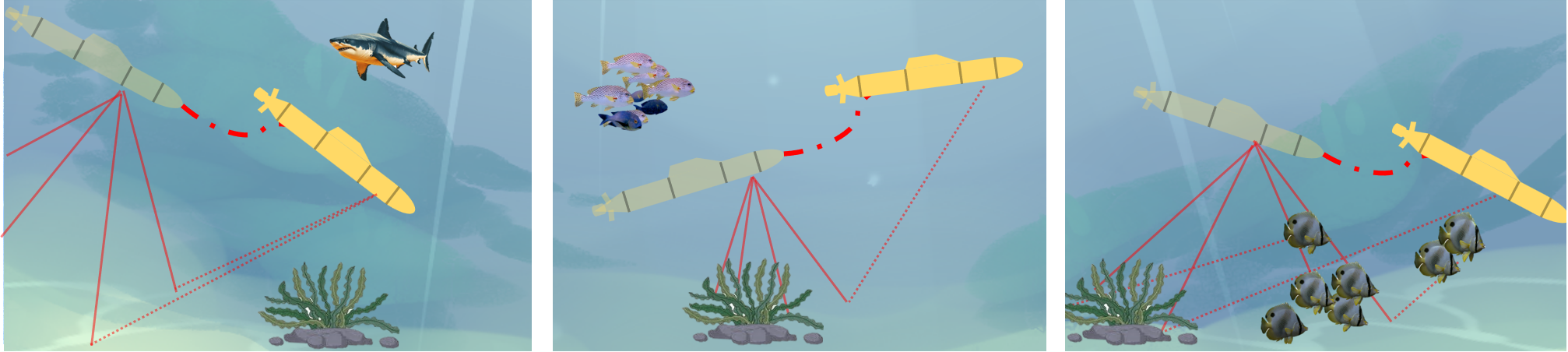} % Adjust width as needed
    \caption{A visualization of various scenarios where an AUV may encounter conditions limiting  DVL beam measurements. On the left-hand side, the initial operation of the AUV, diving, is illustrated, characterized by a significant pitch angle. In the middle, the acoustic beams encounter uneven terrain, while on the right-hand side, the DVL view is obstructed by sea animals.}
    \label{fig:limitedbeams}
\end{figure*}

\IEEEPARstart{N}{avigating} underwater environments has garnered significant attention in recent years due to its importance across scientific missions and various industries, such as oil and gas exploration, marine, subsea, and military operations. Underwater navigation poses challenges because traditional methods reliant on global navigation satellite system (GNSS) signals are ineffective, as these signals cannot penetrate water. The most promising and used sensors for autonomous underwater vehicle (AUV) navigation are the inertial navigation system (INS) and a Doppler velocity log (DVL) \cite{miller2010autonomous}.
\\ \noindent
As a self-contained system, the strapdown INS can offer continuous high-rate updates of the platform's position, velocity, and orientation. The INS derives its navigation solution by integrating the inertial data from the inertial sensors, which are susceptible to inherent errors and noises. These factors contribute to the accumulation of errors in the navigation solution over time \cite{titterton2004strapdown}. To mitigate these errors, an aiding sensor in the form of the DVL is utilized. It operates by transmitting four acoustic beams to the seabed. Upon receiving the reflected beams, the Doppler effect is utilized to calculate the velocity vector of the platform by evaluating the frequency shift between the transmitted and received beams \cite{brokloff1994matrix}. To effectively integrate these sensors, a nonlinear filter is employed due to the nonlinear characteristics of the INS equations of motion. Common choices include the extended Kalman filter (EKF) or the unscented Kalman filter (UKF) \cite{karimi2013comparison}.
\\ \noindent
The DVL operates under a condition known as "bottom lock," which occurs when at least three out of the four transmitted beams are reflected back to the sensor \cite{taudien2018doppler}. When this condition is met, the velocity vector updates provided by the DVL are integrated with the INS using a loosely coupled (LC) approach. This involves initially calculating the velocity vector of the platform through parameter estimation using the beam velocity measurements and the least squares estimator. Subsequently, once the velocity vector is obtained, it is incorporated into the nonlinear filter to mitigate the inertial errors. In real-world scenarios, the bottom lock condition is not always achieved, rendering the LC approach impractical as the LS estimator struggles to estimate Cartesian velocity. To circumvent this challenge, the raw DVL measurements, specifically the beam velocity readings, can be directly integrated into the nonlinear filter using a tightly coupled (TC) approach. This approach is adaptable to any number of beams, ranging from one to four. However, as the number of beams decreases, the system's capability to mitigate navigation errors diminishes \cite{rudolph2012doppler}. 
\\ \noindent
Scenarios involving partial DVL measurements can occur due to various factors, including extreme roll and pitch maneuvers like while diving, passing over underwater structures such as trenches and rocks that may deflect the beams, and obstructions such as marine wildlife blocking the sensor's view. A visualization of these scenarios can be seen in Fig.\ref{fig:limitedbeams}. To tackle these scenarios, research has started to emerge, beginning with Tal \emph{et al.}, who proposed an extended loosely coupled (ELC) approach \cite{tal2017inertial}. This method utilizes partial DVL beam measurements along with external information to compute the three-dimensional velocity vector of the AUV, which is then fed into the navigation nonlinear filter. Additionally, in \cite{liu2018ins}, the authors proposed a TC implementation of INS/DVL along with a pressure sensor to address scenarios involving partial beam measurements. In \cite{yao2022virtual}, the authors analyzed the geometric relationship between the DVL beam configuration and a zero velocity vector assumption. They derived a zero velocity update-aided virtual beam method to continuously operate the TC fusion and also proposed an LS support vector machine-aided virtual beam method to integrate into the TC INS/DVL fusion. Moreover, an approach has been proposed to estimate the platform’s velocity vector based on past DVL measurements and a motion model specifically designed for short time periods \cite{klein2022estimating}. With recent advancements in computational efficiency and proven results of data-driven approaches, deep learning (DL) has started to emerge as a method to address the problems associated with partial beam measurements. In \cite{yona2021compensating}, the authors employed a convolutional neural network (CNN) to regress a single missing beam in cases where it is absent. Subsequently, they investigated various combinations of missing beam regression using a long short-term memory (LSTM) based network in \cite{yona2024missbeamnet}. Our previous work consists of a series of networks referred to as BeamsNet, which demonstrated how one-dimensional CNNs can enhance DVL measurements \cite{cohen2022beamsnet}, compensate for missing beams \cite{cohen2022libeamsnet,cohen2024data}, and provide velocity updates in cases of complete DVL outage \cite{cohen2023set}.
\\ \noindent
In this paper, we leverage our previous work by adapting and modifying the BeamsNet framework to directly regress missing beam measurements instead of the AUV's velocity vector. Subsequently, we integrate the regressed missing beams together with the measured ones into the EKF using both LC and TC integration approaches to form a seamless AUV navigation solution. Specifically, we focus on scenarios involving two and three missing beams. The contributions made in this study are outlined below:
\begin{enumerate}
    \item A tailored BeamsNet architecture, uniquely crafted to forecast missing beams in situations where two or three beams are absent, leveraging both past DVL beam measurements and the current partial measurement.
    \item Hybrid neural LC (HNLC) methodology designed to incorporate incomplete DVL measurements with regressed beam data forming a velocity vector update to the EKF.
    \item Hybrid neural TC (HNTC) methodology fusing regressed and measured beams as an update to the EKF.
\end{enumerate}

\noindent
To emphasize our approach's robustness, we evaluate it using real-world AUV data collected on two separate occasions.
\hlcc{The training and validation datasets consisted of approximately four hours of data, spanning nine distinct missions with varied maneuvers, speeds, depths, and other factors. In contrast, the test dataset comprised a mission of four hundred seconds, conducted on a separate date under different sea conditions. This diverse test scenario aimed to underscore the robustness of the HNC approach.}
In situations of limited beam measurements, we compare HNLC and HNTC to the baseline TC and LC (pure inertial) and a model-based average beam estimator. \hlcc{Our approach offers an improvement of 96.15\% on average over the baseline model-based LC and TC approaches. Additionally, it demonstrates superior performance compared to a model-based beam estimator by an average of 12.41\% in terms of velocity accuracy for scenarios involving two or three missing beams. }\\
\noindent
The rest of the paper is organized as follows: Section \ref{INSDVL} introduces the mathematical background of INS/DVL fusion, examining both the LC and TC approaches. Section \ref{HNC} presents the suggested hybrid neural coupled approach. Finally, Sections \ref{AR} and \ref{Con} discuss the results and the conclusions, respectively.
\section{INS/DVL Fusion}\label{INSDVL}
\noindent
In this section, we describe the implementation of the error-state EKF for INS/DVL fusion. Throughout the implementation, three main reference frames will be indicated with an upper script. The first frame, denoted as $b$, represents the body frame centered at the vehicle's center of mass. In this frame, the x-axis aligns with the vehicle's longitudinal axis, pointing in the forward direction. The z-axis extends downward, and the y-axis extends outward, completing the right-hand orthogonal coordinate system. The second frame, denoted as $n$, represents the navigation frame, which is locally defined relative to the Earth's geoid. We utilize the north-east-down (NED) coordinate system within this frame, where the x-axis aligns with true north, the y-axis points eastward, and the z-axis points downward following the direction of gravity. The final coordinate, denoted as $DVL$, represents the DVL frame, with its sensitive axes determined by the manufacturer. The transformation from the DVL frame to the body frame is described by a fixed transformation matrix, known in advance \cite{farrell2008aided}.
\\ \noindent
The error-state EKF implementation is employed for the INS/DVL fusion. We denote the error-state vector as $\boldsymbol{\delta x} \in \mathbb{R}^{n\times1}$, which represents an $n$-dimensional vector, formulated as:
\begin{equation}\label{eqn:errormodel}
\centering
\boldsymbol{x}^{t}= \boldsymbol{x}^{e}-\boldsymbol{\delta x}
\end{equation}
in this context, $\boldsymbol{x}^{t} \in \mathbb{R}^{n\times1}$ and $\boldsymbol{x}^{e} \in \mathbb{R}^{n\times1}$ denote the true state and the estimated state, respectively. Specifically, in the scenario of INS/DVL fusion, involving $n=12$ error-states, we utilize $\boldsymbol{\delta x^{n}}$:
\begin{equation}\label{eqn:dx}
\centering
\boldsymbol{\delta x}^{T}= [\boldsymbol{(\delta v^{n})}^{T}\quad \boldsymbol{ (\epsilon^{n})}^{T}\quad \boldsymbol{\delta b_{a}}^{T}\quad \boldsymbol{\delta b_{g}}^{T}]^{T} \in \mathbb{R}^{12\times1}
\end{equation}
in this formulation, $\boldsymbol{\delta v^{n}} \in \mathbb{R}^{3\times1}$, $\boldsymbol{\epsilon} \in \mathbb{R}^{3\times1}$, $\boldsymbol{b_{a}} \in \mathbb{R}^{3\times1}$, and $\boldsymbol{b_{g}} \in \mathbb{R}^{3\times1}$ denote the velocity error-states expressed in the navigation frame, misalignment error, accelerometer bias residual error, and gyroscope bias residual error, respectively \cite{rogers2003applied}. The linearized differential equation governing the error state is represented as follows:
\begin{equation}\label{eqn:f}
\centering
\boldsymbol{\delta \dot{x}}= \mathbf{F}\boldsymbol{\delta x} +\mathbf{G}\boldsymbol{n}
\end{equation}
where $\boldsymbol{n}\in \mathbb{R}^{12\times1}$ denotes the system noise vector, $\mathbf{F}\in \mathbb{R}^{12\times12}$ represents the system matrix, and $\mathbf{G}\in \mathbb{R}^{12\times12}$ is the system noise distribution matrix. The system incorporates multiple independent sources of noise, each assumed to follow a zero-mean Gaussian distribution. These sources of noise can be represented as:
\begin{equation}\label{eqn:noise}
\centering
\boldsymbol{n}=[\boldsymbol{n_{a}}^{T}\quad \boldsymbol{ n_{g}}^{T}\quad \boldsymbol{n_{a_{b}}}^{T}\quad \boldsymbol{n_{g_{b}}}^{T}]^{T}\in \mathbb{R}^{12\times1}.
\end{equation}
where $\boldsymbol{n_{a}}\in\mathbb{R}^{3\times1}$ and $\boldsymbol{n_{g}}\in\mathbb{R}^{3\times1}$ represent the additive noise for the accelerometer and gyroscope, respectively. Additionally, $\boldsymbol{n_{a_{b}}}\in\mathbb{R}^{3\times1}$ and $\boldsymbol{n_{g_{b}}}\in\mathbb{R}^{3\times1}$ denote the Gaussian random walk distributions for the accelerometer and gyroscope, respectively.
The system matrix, $\mathbf{F}$, is defined as:
\begin{equation}\label{eqn:SM}
    \centering 
    \mathbf{F}=
        \begin{bmatrix} 
        \mathbf{F}_{vv}&
        \mathbf{F}_{v\epsilon}&
        \mathbf{0}_{3\times3}&
        \mathbf{C}_{b}^{n}\\
        \mathbf{F}_{\epsilon v}&
        \mathbf{F}_{\epsilon\epsilon}&
        \mathbf{C}_{b}^{n}&
        \mathbf{0}_{3\times3}\\
        \mathbf{0}_{3\times3}&
        \mathbf{0}_{3\times3}&
        \mathbf{0}_{3\times3}&
        \mathbf{0}_{3\times3}\\
        \mathbf{0}_{3\times3}&
        \mathbf{0}_{3\times3}&
        \mathbf{0}_{3\times3}&
        \mathbf{0}_{3\times3}\\
    \end{bmatrix}
\end{equation} 
where $\mathbf{C}_{b}^{n}$ denotes the transformation matrix from the body frame to the navigation frame. The specific sub-matrices of $\mathbf{F}$ can be referenced in textbooks like \cite{savage2000strapdown,groves2015principles}.
Concerning the distribution of system noise, the matrix $\mathbf{G}$ is defined as follows:
\begin{equation}\label{eqn:SG}
\centering
\mathbf{G}=
\begin{bmatrix}
\mathbf{0}_{3\times3}&
\mathbf{C}_{b}^{n}&
\mathbf{0}_{3\times3}&
\mathbf{0}_{3\times3}\\
\mathbf{C}_{b}^{n}&
\mathbf{0}_{3\times3}&
\mathbf{0}_{3\times3}&
\mathbf{0}_{3\times3}\\
\mathbf{0}_{3\times3}&
\mathbf{0}_{3\times3}&
\mathbf{I}_{3\times3}&
\mathbf{0}_{3\times3}\\
\mathbf{0}_{3\times3}&
\mathbf{0}_{3\times3}&
\mathbf{0}_{3\times3}&
\mathbf{I}_{3\times3}\\
\end{bmatrix}
\end{equation}
where $\mathbf{0}_{3\times3}$ represents a three-by-three zero matrix, and $\mathbf{I}_{3\times3}$ denotes a three-dimensional identity matrix.
\\ \noindent
The Kalman filtering process typically consists of two distinct phases: the prediction and update steps. During the prediction step, the \textit{a priori} error state, denoted as $\delta\boldsymbol{x}^{-}$, is initialized to zero to facilitate linearization, which is a crucial aspect of the EKF method.
\begin{equation}\label{eqn:apriorizero}
\centering
\delta\boldsymbol{x}^{-}=0
\end{equation}
Next, the state covariance is propagated using the known model, which in this case corresponds to the INS nonlinear equations of motion:
\begin{equation}\label{eqn:pred}
\centering
\mathbf{P}^{-}_{k}=\mathbf{\Phi}_{k-1}\mathbf{P}^{+}_{k-1}\mathbf{\Phi}_{k-1}^{T}+\mathbf{Q}_{k-1}
\end{equation}
Here, $\mathbf{P}^{-}_{k}$ denotes the \textit{a priori} state covariance estimate at time $k$, and $\mathbf{P}^{+}_{k-1}$ represents the \textit{a posteriori} state covariance estimate at time $k-1$. The transition matrix, denoted as $\mathbf{\Phi}_{k-1}$, is typically derived through a power-series expansion of the system matrix, $\mathbf{F}$, and the propagation interval, $\tau_{s}$:
\begin{equation}\label{eqn:transition}
\centering
     \mathbf{\Phi}_{k-1} = \sum_{r=0}^{\infty} \frac{\mathbf{F}^{r}_{k-1}}{r!} \tau_{s}^{r}
\end{equation}
The discrete process noise covariance matrix is derived from its continuous form $\mathbf{Q} = \mathbb{E}[\boldsymbol{n}\boldsymbol{n}^{T}]$. It accounts for the inherent uncertainty in the model and is typically approximated as follows: \cite{gelb1974applied}:
\begin{equation}\label{eqn:Q_dis_MB}
\mathbf{Q}_{k-1} = \frac{1}{2} \left(\mathbf{\Phi}_{k-1} \mathbf{G}_{k-1} \mathbf{Q} \mathbf{G}_{k-1}^T + \mathbf{G}_{k-1} \mathbf{Q} \mathbf{G}_{k-1}^T \mathbf{\Phi}_{k-1}^T\right)  \Delta t.
\end{equation}
\noindent
The next stage of the Kalman filter is the update step, which is executed using the following equations:
\begin{equation}\label{gain}
\mathbf{K}_{k} = \mathbf{P}^{-}_{k}\mathbf{H}^{T}_{k}\left(\mathbf{H}_{k}\mathbf{P}^{-}_{k}\mathbf{H}^{T}_{k} +\mathbf{R}_{k} \right)^{-1}
\end{equation}
\begin{equation}\label{postriori}
\mathbf{P}^{+}_{k} =[\mathbf{I}-\mathbf{K}_{k}\mathbf{H}_{k} ] \mathbf{P}^{-}_{k}
\end{equation}
\begin{equation}\label{dz}
\boldsymbol{\delta{x}}^{+}_{k}=\mathbf{K}_{k}\boldsymbol{\delta{z}}_{k}
\end{equation}
where $\mathbf{K}_{k}$ represents the Kalman gain, which balances between incorporating new measurements and predictions from the system's dynamic model. The matrices $\mathbf{H}_{k}$ and $\mathbf{R}_{k}$ correspond to the measurement matrix and noise covariance matrix, respectively. Finally, $\boldsymbol{\delta{x}}^{+}_{k}$ and $\mathbf{P}^{+}_{k}$ denote the \textit{a posteriori} error state and the estimated covariance state, respectively.
\\ 
\noindent
To determine the measurement matrix $\mathbf{H}_{k}$, it is essential to first inspect the geometry of the DVL operation. The DVL is installed inside the AUV such that its transducers are oriented toward the seabed. The DVL operates in an "$\times$" type configuration, also known as the "Janus Doppler configuration" \cite{cohen2024kit}. In the "$\times$" type configuration, the transactors are located with a relative yaw and pitch angles to the DVL's sensor body frame, which usually differs from the platform's body frame, in the following manner:
\begin{equation}\label{eqn:4}
    \centering
        \boldsymbol{b}_{\dot{\imath}}=
        \begin{bmatrix} 
        \cos{\psi_{\dot{\imath}}}\sin{\theta}\quad
        \sin{\psi_{\dot{\imath}}}\sin{\theta}\quad
        \cos{\theta}
    \end{bmatrix}_{1\times3}
\end{equation} 
where $\boldsymbol{b}_{\dot{\imath}}$ such that $\dot{\imath}=1,2,3,4$ represents the beam number, and $\psi$ and $\theta$ denote the yaw and pitch angles relative to the body frame, respectively. The pitch angle is fixed and predetermined by the manufacturer, maintaining the same value for each beam. The yaw angle can be expressed, for example, by \cite{tal2017inertial}:
\begin{equation}\label{eqn:5}
    \centering
        \psi_{\dot{\imath}}=(\dot{\imath}-1)\cdot90^{\circ}+45^{\circ}\;,\; \dot{\imath}=1,2,3,4
\end{equation}
\subsection{Loosely Coupled INS/DVL Fusion}
\noindent
In the LC approach, the AUV velocity vector estimation through raw DVL beam measurements, is conducted separately to the EKF framework, utilizing a parameter estimation method known as least squares (LS) parameter estimation. Let $\boldsymbol{v}^{DVL}$ represent the velocity vector expressed in the DVL frame, and $\mathbf{T}$ be the transformation matrix mapping it to the velocity in the beam directions, denoted by $\boldsymbol{v}^{Beam}$. Therefore, the relation between them can be defined as follows:
\begin{equation}\label{eqn:6}
    \centering
        \boldsymbol{v}^{Beam}=\mathbf{T}\boldsymbol{v}^{DVL} ,\quad
        \mathbf{T}=
        \begin{bmatrix} \boldsymbol{b}_{1}\\\boldsymbol{b}_{2}\\\boldsymbol{b}_{3}\\\boldsymbol{b}_{4}\\
    \end{bmatrix}_{4\times3}
\end{equation} 
The beam measurements are subject to inherent errors, which are modeled by:
\begin{equation}\label{eqn:77}
    \centering
        \boldsymbol{y}= \mathbf{T}[\boldsymbol{v}^{DVL}(\boldsymbol{1}+\boldsymbol{s}_{DVL})]+\boldsymbol{b}_{DVL}+\boldsymbol{n}_{DVL}
\end{equation}
where, $\boldsymbol{b}_{DVL}\in \mathbb{R}^{4\times1}$ represents the bias vector, $\boldsymbol{s}_{DVL}\in \mathbb{R}^{3\times1}$ denotes the scale factor vector, and $\boldsymbol{n}_{DVL}\in \mathbb{R}^{4\times1}$ stands for a zero-mean white Gaussian noise. After obtaining the raw measurements, the subsequent step involves extracting $\boldsymbol{v}^{d}$ by filtering the data based on the following cost function:
\begin{equation}\label{eqn:8}
    \centering
        \hat{\boldsymbol{v}}^{DVL}=
        \underset{\boldsymbol{v}^{DVL}}{\argmin}{\mid\mid\boldsymbol{y}-\mathbf{T}\boldsymbol{v}^{DVL} \mid\mid}^{2}.
\end{equation} 
The solution to this LS problem is obtained by multiplying the observations with the pseudo-inverse of the matrix $\mathbf{T}$ \cite{bar2004estimation}:
\begin{equation}\label{eqn:99}
    \centering
        \hat{\boldsymbol{v}}^{DVL}=(\mathbf{T}^{T}\mathbf{T})^{-1}\mathbf{T}^{T}\boldsymbol{y}.
\end{equation}
Finally, the estimated velocity vector from the DVL is transformed to the body frame using:
\begin{equation}\label{eqn:199}
    \centering        \hat{\boldsymbol{v}}^{b}_{DVL}=\mathbf{C}_{d}^{b}\hat{\boldsymbol{v}}^{DVL}
\end{equation}
where $\mathbf{C}_{d}^{b}$ represents a known constant transformation matrix from the DVL frame to the body frame and $\hat{\boldsymbol{v}}^{b}_{DVL}$ denotes the DVL velocity in the body frame.
\\ \noindent
To derive $\mathbf{H}_{k}$ for the LC approach, we examine the innovation residual and employ a first-order perturbation approximation while neglecting higher-order error components:
\begin{equation}\label{eqn:dz}
\begin{split}
    \boldsymbol{\delta{z}} & = \hat{\mathbf{C}}_{n}^{b}\hat{\boldsymbol{v}}^{n}-\boldsymbol{v}^{b}_{DVL} \\
    & =\mathbf{C}_{n}^{b}(\mathbf{I}_{3}+\boldsymbol{\epsilon} [\times])(\boldsymbol{v}^{n}+\delta \boldsymbol{v}^{n}) - \boldsymbol{v}^{b}_{DVL} \\
    & = \cancelto{\boldsymbol{v}^{b}_{DVL} }{\mathbf{C}_{n}^{b}\boldsymbol{v}^{n}}+\mathbf{C}_{n}^{b}\delta \boldsymbol{v}^{n} + \mathbf{C}_{n}^{b}\boldsymbol{\epsilon} [\times]\boldsymbol{v}^{n} -\cancelto{0}{\mathbf{C}_{n}^{b}\boldsymbol{\epsilon} [\times]\delta \boldsymbol{v}^{n}} - \boldsymbol{v}^{b}_{DVL} \\
    & \approx \mathbf{C}_{n}^{b}\delta \boldsymbol{v}^{n} - \mathbf{C}_{n}^{b}\boldsymbol{v}^{n}[\times] \boldsymbol{\epsilon}
\end{split}
\end{equation}
therefore $\mathbf{H}_{k}$ can be formulated into the following:
\begin{equation}\label{H_LC}
\centering
    \mathbf{H}_{k}^{LC}=[\mathbf{C}_{n}^{b} \quad -\mathbf{C}_{n}^{b}\boldsymbol{v}^{n}[\times] \quad \mathbf{0}_{3\times3}\quad \mathbf{0}_{3\times3}]
\end{equation}
The loosely coupled integration offers two primary advantages: simplicity and redundancy. However, its main drawbacks arise from the use of cascaded filters (LS and EKF), which results in a sub-optimal solution, and the requirement of at least three available beams to estimate the velocity vector.
\subsection{Tightly Coupled INS/DVL Fusion}
\noindent
In the TC approach, the DVL raw beam measurements are directly processed through the EKF update stage. Each raw beam measurement is handled separately in the following manner:
\begin{equation}\label{eqn:dBeam}
\begin{split}
    \boldsymbol{\delta{z}}_{i} & = \boldsymbol{b}_{i}^{T}\hat{\mathbf{C}}_{n}^{b}\hat{\boldsymbol{v}}^{n}-\boldsymbol{b}_{i}^{T}\boldsymbol{v}^{b}_{DVL} \\
    & =\boldsymbol{b}_{i}^{T}\mathbf{C}_{n}^{b}(\mathbf{I}_{3}+\boldsymbol{\epsilon} [\times])(\boldsymbol{v}^{n}+\delta \boldsymbol{v}^{n}) - \boldsymbol{b}_{i}^{T}\boldsymbol{v}^{b}_{DVL} \\
    & = \cancelto{\boldsymbol{b}_{i}^{T}\boldsymbol{v}^{b}_{DVL} }{\boldsymbol{b}_{i}^{T}\mathbf{C}_{n}^{b}\boldsymbol{v}^{n}}+\boldsymbol{b}_{i}^{T}\mathbf{C}_{n}^{b}\delta \boldsymbol{v}^{n} + \boldsymbol{b}_{i}^{T}\mathbf{C}_{n}^{b}\boldsymbol{\epsilon} [\times]\boldsymbol{v}^{n}\\
    & - \cancelto{0}{\boldsymbol{b}_{i}^{T}\mathbf{C}_{n}^{b}\boldsymbol{\epsilon} [\times]\delta \boldsymbol{v}^{n}} - \boldsymbol{b}_{i}^{T}\boldsymbol{v}^{b}_{DVL} \\
    & \approx \boldsymbol{b}_{i}^{T}\mathbf{C}_{n}^{b}\delta \boldsymbol{v}^{n} - \boldsymbol{b}_{i}^{T}\mathbf{C}_{n}^{b}\boldsymbol{v}^{n}[\times] \boldsymbol{\epsilon}
\end{split}
\end{equation}
where $\boldsymbol{\delta{z}}_{i}$ is the residual of the $i^{th}$ beam measurement and therefore $\mathbf{H}_{k,i}$, that corresponds to this beam, can be formulated into the following:
\begin{equation}\label{H_TC}
\centering
    \mathbf{H}_{k,i}^{TC}=[\boldsymbol{b}_{i}^{T}\mathbf{C}_{n}^{b} \quad -\boldsymbol{b}_{i}^{T}\mathbf{C}_{n}^{b}\boldsymbol{v}^{n}[\times] \quad \mathbf{0}_{1\times3}\quad \mathbf{0}_{1\times3}]
\end{equation}
One of the advantages of the TC approach is its ability to operate in cases where there is no bottom lock, meaning that at least three beams are reflected back to the sensor from the seabed, including scenarios with fewer than three beams. The actual $\mathbf{H}_{k}$ is determined by the number of available beams by concatenating the $\mathbf{H}_{k,i}$. For example, in an ideal case with four beams, the measurement matrix would be:
\begin{equation}\label{H_TC4}
\centering
        \mathbf{H}_{k}^{TC}=
        \begin{bmatrix} 
            \mathbf{H}_{k,1}^{TC}\\\mathbf{H}_{k,2}^{TC}\\\mathbf{H}_{k,3}^{TC}\\\mathbf{H}_{k,4}^{TC}\\
    \end{bmatrix}_{4\times12}
\end{equation}
It has been demonstrated in the literature that the TC approach consistently outperforms its LC counterpart in terms of both accuracy and robustness \cite{zhang2023autonomous}.
\section{Hybrid Neural Coupled INS/DVL Fusion}\label{HNC}
\noindent
In this section, we introduce a deep learning methodology inspired by the BeamsNet framework \cite{cohen2022beamsnet,cohen2022libeamsnet} to allow regression of missing beams and thus enable seamless AUV navigation in scenarios with limited DVL measurements. Additionally, we explore the feasibility of integrating our hybrid neural approach into the error-state EKF framework using both LC and TC approaches.
\subsection{Missing Beams Regression}
\begin{figure}[h!] % [t] option to align at the top
    \centering
    \includegraphics[width=\columnwidth]{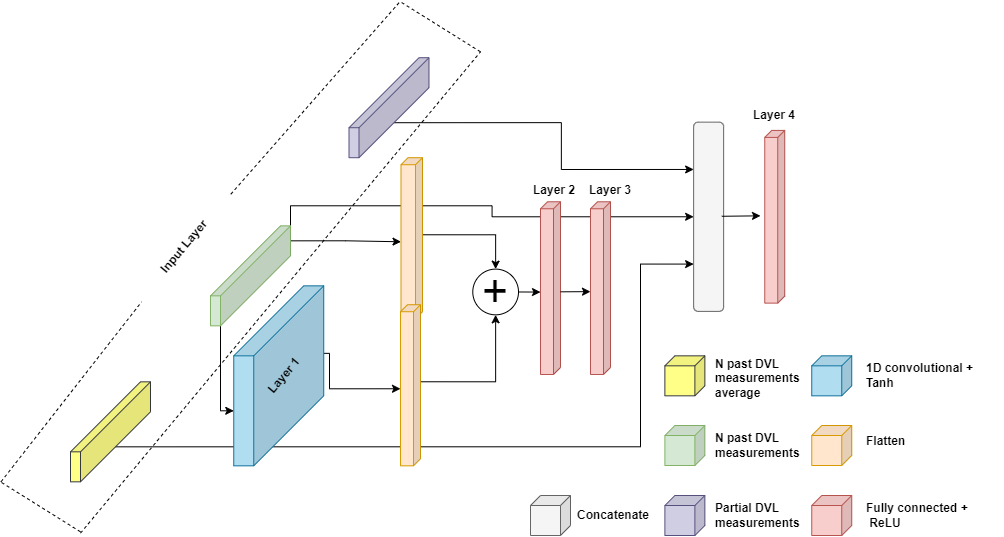} % Adjust width as needed
    \caption{A schematic representation of the revised BeamsNet architecture, comprising of one-dimensional convolutional layers and fully connected layers.}
    \label{fig:arc}
\end{figure}

\noindent
Our proposed framework utilizes $N$ past DVL measurements and the current partial beam measurements (one or two). Considering that AUVs generally do not engage in extreme maneuvers and commonly operate at speeds lower than 4 [m/s], we opted to concentrate on a brief time window $N$ preceding the partial beam measurements. In doing so, we leverage a data-driven methodology to extract pertinent features facilitating the prediction of missing beams. The proposed approach's architecture relies on a one-dimensional convolutional neural network (1DCNN), as depicted in Fig. \ref{fig:arc}.
The first step of the network is feature extraction from past DVL measurements using a 1D convolutional layer with a kernel size of $2\times1$ and stride of $2\times1$, followed by a hyperbolic tangent function. Then, a residual connection between the flattened output of this layer and the flattened input is established and inserted into a set of fully connected layers followed by rectified linear units. The output of this section is concatenated with the current partial beam measurements, along with an additional input, which is the average over the $N$ past DVL measurements. Lastly, it passes through a fully connected layer that outputs the missing beams.
The network can also be formulated mathematically in the following manner:
\begin{subequations}
\begin{equation}\label{X_input}
    \text{Input:} \quad X = [x_1, x_2, \ldots, x_N]
\end{equation}
\begin{equation}\label{P_input}
    \text{Input:} \quad P
\end{equation}
\end{subequations}
where \( X \) is a vector of \( N \) past DVL beam measurements and \( P \) represents the current partial DVL beam measurements. \( M \) is the third input representing the mean of the past DVL beam measurements:
\begin{equation}\label{M_mean}
    M = \frac{1}{N} \sum_{i=1}^{N} x_i
\end{equation}
In the \(i^{th}\) convolutional layer, the output \( Z_{i} \) is obtained as follows:
\begin{equation}\label{Zi}
    Z_{i} = \sigma_{\text{Tanh}}\left(\sum_{j=1}^{m} (x_{i+(j-1)s} \cdot w_j) + b\right)
\end{equation}
where  \( m \) \hlcc{represents the window/kernel size}, \( b \) \hlcc{signifies the bias}, \( s \) \hlcc{denotes the stride}, and \( w_j \) \hlcc{represents the weights}. \hlcc{Additionally,} $\sigma_{\text{Tanh}}$ \hlcc{denotes the Tanh activation function, defined as follows:}
\begin{equation}
    \sigma_{\text{Tanh}}(x) = \frac{{e^{x} - e^{-x}}}{{e^{x} + e^{-x}}}
\end{equation}
Combining the output of the flattened convolutional layer \( Z \) \eqref{Zi} with the input \( X \) \eqref{X_input}, resulting in layer \( Y \):
\begin{equation}
    Y = Z + X
\end{equation}
The output \( L \) of the first fully connected layer is computed as:
\begin{equation}\label{L}
    L = \sigma_{\text{ReLU}}(W_2 ( \sigma(W_1 Y + b_1) ) + b_2)
\end{equation}
\hlcc{where}  \( W_1 \) and \( W_2 \) \hlcc{represent the weight matrices}, and \( b_1 \) and \( b_2 \) \hlcc{ represent the bias vectors.} \hlcc{Additionally,} $ \sigma_{\text{ReLU}}$ \hlcc{denotes the ReLU activation function, defined as follows:}
\begin{equation}
   \sigma_{\text{ReLU}}(x) = \max(0, x)
\end{equation}
After obtaining \( L \) in \eqref{L},  it is stacked with \( P \) \eqref{P_input} and \( M \) \eqref{M_mean} to form \( U \):
\begin{equation}
    U = \text{stack}\{ L,P,M\}
\end{equation}
Finally, the output \( \hat{O} \) of the neural network is computed as:
\begin{equation}\label{DNN_out}
    \hat{O} = \sigma_{\text{ReLU}}(W_3 U + b_3)
\end{equation}
\hlcc{where}  \( W_3 \) \hlcc{is the weight matrix} and \( b_3 \) \hlcc{ is the bias vector.}
During the training process, we employ the mean squared error (MSE) loss function:
\begin{equation}
    \text{MSE}(O,\hat{O}) = \frac{1}{n} \sum_{i=1}^{n} (O_i - \hat{O}_i)^2
\end{equation}
where \({O} \) represents the reference data and $n$ in the number of data points. 
\\ 
\noindent
The optimization process is carried out using the root mean square propagation (RMSprop) optimizer. The model is trained with a batch size of 4, a learning rate of 0.001, and a learning rate decay factor of 0.1 every 35 epochs. Training is conducted over 100 epochs. The differences in hyper-parameters between the architectures for handling two or three missing beams are presented in Table \ref{tab:param}. Notice that the same architecture is applied with differences only in the input and output shapes.
\begin{table}[h!]
\centering
\caption{Variations in hyperparameters for two and three missing beam networks within the modified BeamsNet architecture.}
\resizebox{\columnwidth}{!}{%
\begin{tabular}{|c|c|c|c|}
\hline
\textbf{Network} & \textbf{Type} & \textbf{Input Shape} & \textbf{Output Shape} \\ \hline
\multirow{4}{*}{2 Missing Beams} & Layer1 - Conv1D & 3 & 6 \\ \cline{2-4} 
 & Layer2 - FC & 12 & 16 \\ \cline{2-4} 
 & Layer3 - FC & 16 & 2 \\ \cline{2-4}
 & Layer4 - FC & 8 & 2 \\ \hline
\multirow{4}{*}{3 Missing Beams} & Layer1 - Conv1D & 5 & 10 \\ \cline{2-4} 
 & Layer2 - FC & 20 & 16 \\ \cline{2-4} 
 & Layer3 - FC & 16 & 3 \\ \cline{2-4}
 & Layer4 - FC & 8 & 3 \\ \hline
\end{tabular}%
}
\label{tab:param}
\end{table}
\subsection{Hybrid Neural Coupled EKF}
\noindent
\begin{figure}[h!] % [t] option to align at the top
    \centering
    \includegraphics[width=\columnwidth]{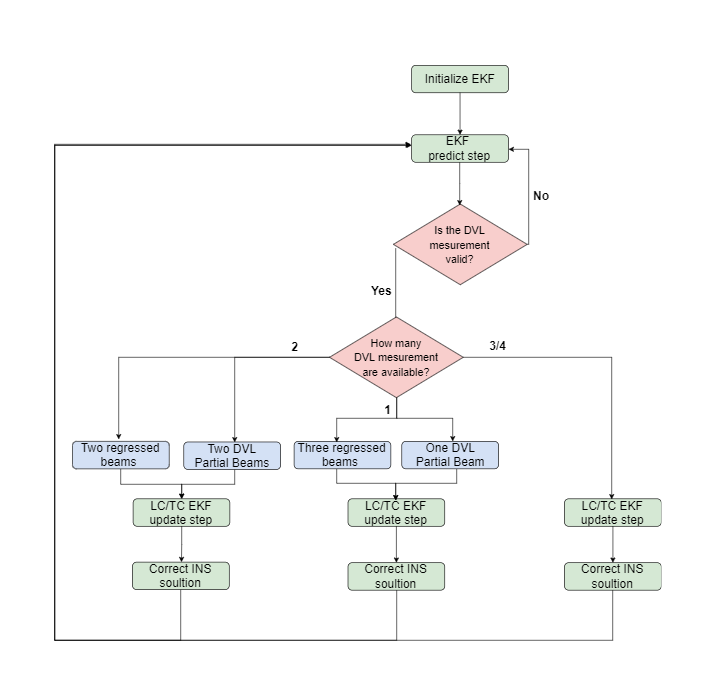} % Adjust width as needed
    \caption{Hybrid-Neural EKF information flow. The EKF is initialized and constantly propagates the INS model while awaiting the DVL velocity update. Upon receiving the DVL velocity update, if the bottom lock condition is met, the update procedure is carried out regularly, either in an LC or TC approach. However, if there are fewer than three beams, the data is directed through the appropriate path and the missing beams are forecasted using the proposed approach and subsequently utilized within the filter.}
    \label{fig:flow}
\end{figure}
Once the network is trained, it is incorporated into the navigation filter to address scenarios of partial beam measurements. When the system identifies the number of missing beams,  \eqref{eqn:77} undergoes a slight modification. Now, the vector $\boldsymbol{y}$ depicted in \eqref{eqn:77} is no longer in $\mathbb{R}^{4\times1}$ but rather in either $\mathbb{R}^{2\times1}$ or $\mathbb{R}^{1\times1}$, depending on the number of  beam measurements. The output of the suggested approach completes the missing beams to obtain a total of four beam measurements. Let the beams be denoted as beam1 to beam4 according to \eqref{eqn:4}. For the two missing beams scenario,  we arbitrarily choose beam1 and beam3 as the missing beams, while when three beams are missing, we arbitrarily choose beam2 to be available. Therefore, for the scenario of two missing beams, the measurement vector is defined as:
\begin{equation}
\boldsymbol{y}^{HNC}_{2beams} = [\hat{O}[1],\quad \boldsymbol{y}[1],\quad \hat{O}[2],\quad \boldsymbol{y}[2] ] \in \mathbb{R}^{4\times1}
\end{equation}
 and for three missing beams as
\begin{equation}
\boldsymbol{y}^{HNC}_{3beams} = [\hat{O}[1],\quad \boldsymbol{y}[1],\quad \hat{O}[2],\quad \hat{O}[3] ]\in \mathbb{R}^{4\times1}
\end{equation}
where $\boldsymbol{y}[n]$ and $\hat{O}[n]$ represent the beam measurements in the $n^{th}$ position of the DVL and the network's output as defined in \eqref{DNN_out}, respectively. These measurements can then be integrated into the filter using the HNLC approach or the HNTC approach. Notice that in both cases, the measurement matrix remains consistent with that of a regular situation involving complete four-beam measurements, even though the missing beams are obtained from a different source (network regression). This ensures the preservation of the filter update structure. Specifically, $\mathbf{H}_{k}^{HNLC}$ is equivalent to $\mathbf{H}_{k}^{LC}$ as defined in \eqref{H_LC}, and $\mathbf{H}_{k}^{HNTC}$ is equivalent to $\mathbf{H}_{k}^{TC}$ as defined in \eqref{H_TC}.
\\ \noindent
The flow of information, as outlined in Fig.\ref{fig:flow}, begins by propagating the navigation solution using the INS model and its corresponding estimated covariance as per \eqref{eqn:pred}.
Then, when a DVL sample is deemed valid, the first step is to determine the number of available beams. If there are three or four beams available, the update phase proceeds as usual, following either an LC or TC integration approach. If fewer beams are available, the previously stored past DVL beam measurements, along with the current partial measurements, are inputted into our proposed deep learning model to forecast the missing beams. Once all four beams are available, they are used to update the filter either by the LC or TC methods. 
\section{Analysis and Results}\label{AR}
\subsection{Field Experiments and Dataset}
\begin{figure}[h!] % [t] option to align at the top
    \centering
    \includegraphics[width=\columnwidth]{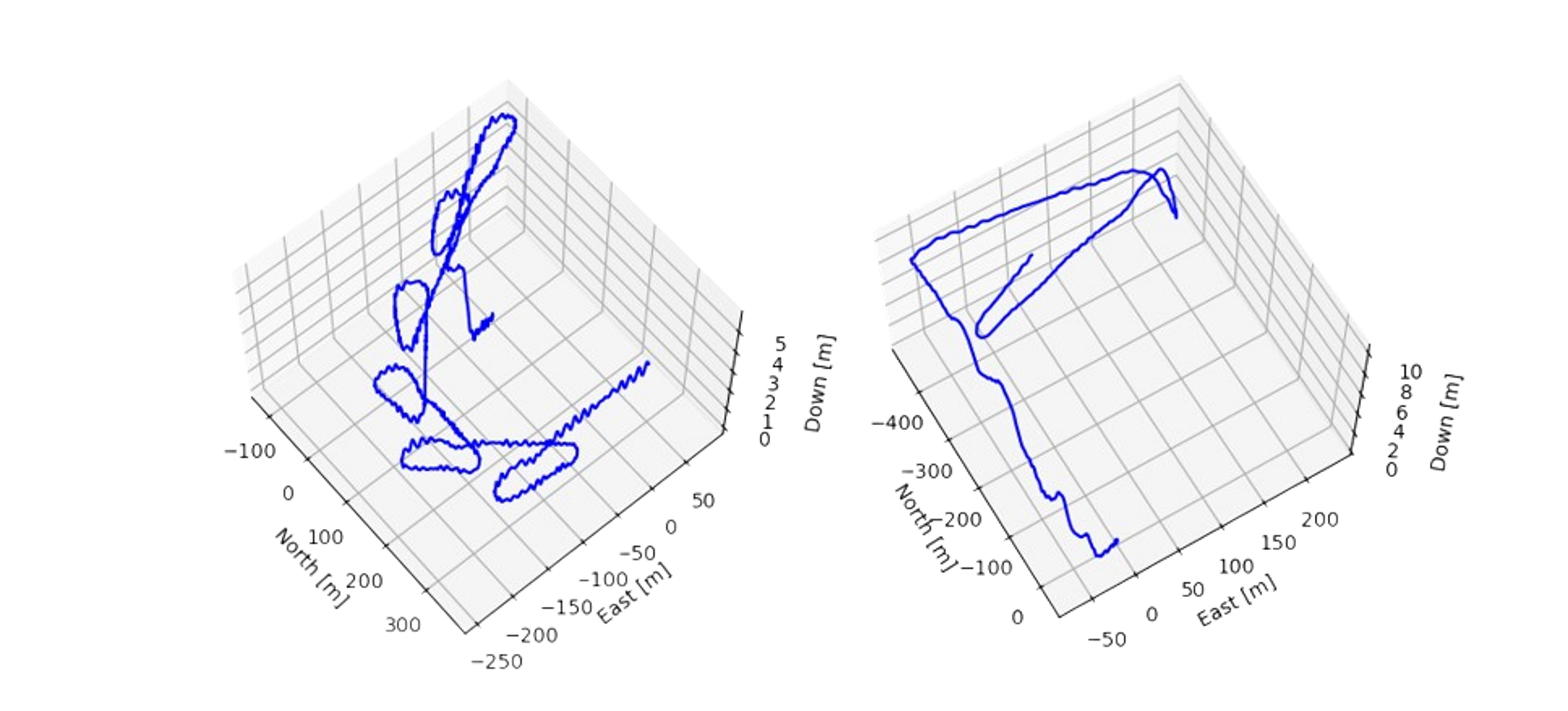} % Adjust width as needed
    \caption{Example of two out of the eight trajectories used to train our proposed network. }
    \label{fig:train_traj}
\end{figure}

\noindent
To evaluate our approach, we generated data by conducting missions with an AUV in the Mediterranean Sea. Specifically, we employed a modified ECA Group A18D mid-size AUV called Snapir. It is capable of autonomously performing missions up to 3000 meters in depth with 21 hours of endurance \cite{ECA_AUV}. The Snapir AUV is outfitted with the iXblue Phins Subsea INS, which utilizes fiber optic gyroscope (FOG) technology for precise inertial navigation \cite{iXblue_PHINS}. Additionally, Snapir uses a Teledyne RDI Work Horse Navigator DVL \cite{TeledyneMarine_DVL}, renowned for its capability to provide accurate velocity measurements with a standard deviation of $0.02\;[m/s]$. The INS operates at a frequency of $100\;[\text{Hz}]$, while the DVL operates at $1\;[\text{Hz}]$.
\\ \noindent
To thoroughly assess the robustness of our approach, we utilized data from two different dates. One dataset was employed for training and validating our data-driven approach, while the testing data comes from a separate mission. The dataset utilized for training and validation was collected in May 2021. The dataset comprises nine distinct missions characterized by various parameters, including maneuvers, depth, speed, sea conditions, and more. Eight of the missions were utilized for training, containing 12,799 DVL samples, equivalent to 3.5 hours of data. An example of two trajectories can be seen in Fig. \ref{fig:train_traj}. Additionally, 1758 DVL samples were reserved for validation, which we refer to as trajectory 1. For testing, an additional 400 seconds of data from another sea experiment conducted in June 2022 were employed.  This data is denoted as trajectory 2. Additionally, corresponding inertial measurement unit (IMU) measurements were captured and utilized in the EKF to propagate the INS and filter model. The ground truth (GT) solution is the one provided by the Delph INS filter using a post-processing software designed for iXblue’s INS-based subsea navigation \cite{iXblue_DELPHINS}.
\\ \noindent
The data provided by the Teledyne RDI Work Horse Navigator DVL comprises velocity samples. To obtain the beam velocity measurements, we utilized the error model specified in \eqref{eqn:77}, considering a DVL bias of 0.01 [m/s] and additive white noise with a standard deviation of 0.042 [m/s] in all three axes and for both missing beam cases. The angle $\theta$ in \eqref{eqn:4} was set to $20^{\circ}$, and the angle $\psi$, defined in \eqref{eqn:5}, is utilized to construct the matrix $\mathbf{T}$ as in \eqref{eqn:6}.
\subsection{Evaluation Metrics}
\noindent
To evaluate the effectiveness of our proposed method, we utilized the root mean square error (RMSE) metric, specifically looking at the velocity RMSE (VRMSE):
\begin{equation}\label{eqn:VRMSE}
\centering
VRMSE(\boldsymbol{v}_{\dot\imath},\hat{\boldsymbol{v}}_{\dot\imath})=\sqrt{\frac{\sum_{\dot\imath=1}^{M}(\boldsymbol{v}_{\dot\imath}-\hat{\boldsymbol{v}}_{\dot\imath})^{2}}{M}}
\end{equation}
where, $M$ represents the total number of samples, $\boldsymbol{v}_{\dot{\imath}}$ represents the ground truth velocity vector, while $\hat{\boldsymbol{v}}_{\dot{\imath}}$ signifies the predicted velocity vector.
\\ \noindent
Additionally, the velocity relative total error (VRTE) is utilized as a percentage to gauge the relative discrepancy between the model's predictions and the reference data using the VRMSE metric. The formula computes the absolute difference between the model's and GT VRMSE, and  then normalized by the GT VRMSE. Finally, it's multiplied by 100 to express the error in percentage.
\begin{equation}\label{eqn:RTE}
\centering
VRTE = \left( \frac{| \text{VRMSE}_{\text{model}} - \text{VRMSE}_{\text{reference}} |}{\text{VRMSE}_{\text{reference}}} \right) \times 100
\end{equation}
where $\text{VRMSE}_{\text{model}}$ and $\text{VRMSE}_{\text{reference}}$ denote the VRMSE of the model's predictions and the GT data, respectively.
\subsection{Experimental Results}
\begin{figure}[h!]
  \centering
  \begin{subfigure}[b]{\columnwidth}
    \includegraphics[width=\columnwidth]{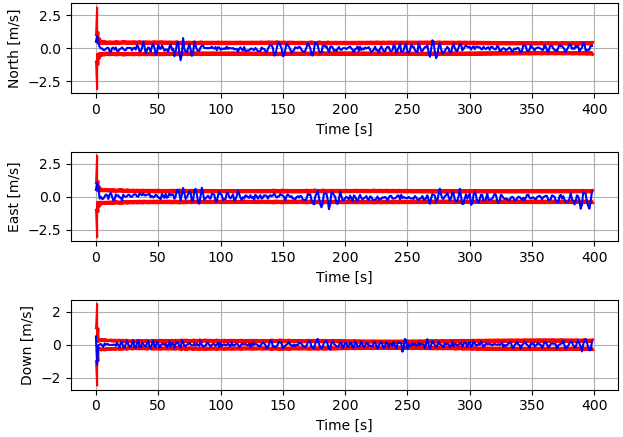}
    \caption{Results of trajectory 1.}
    \label{fig:sub1}
  \end{subfigure}
  \hfill
  \begin{subfigure}[b]{\columnwidth}
    \includegraphics[width=\columnwidth]{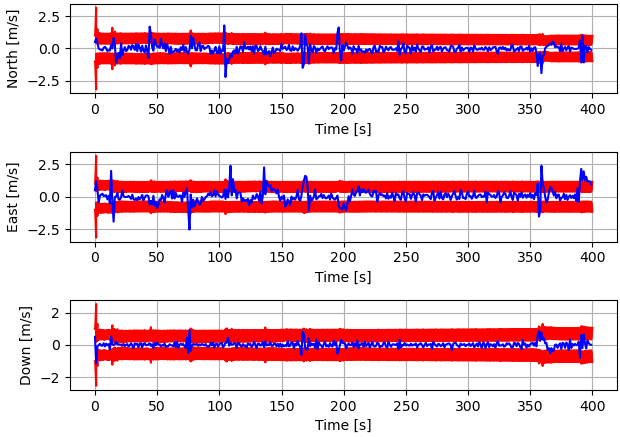}
    \caption{Results of trajectory 2.}
    \label{fig:sub2}
  \end{subfigure}
  \caption{Velocity error states. The blue line represents the velocity error state vector bounded within the estimated covariance sleeve (shown in red) of the EKF for trajectory 1 in (a) and trajectory 2 in (b).}
  \label{fig:velSleeve}
\end{figure}

\begin{figure}[h!]
  \centering
  \begin{subfigure}[b]{\columnwidth}
    \includegraphics[width=\columnwidth]{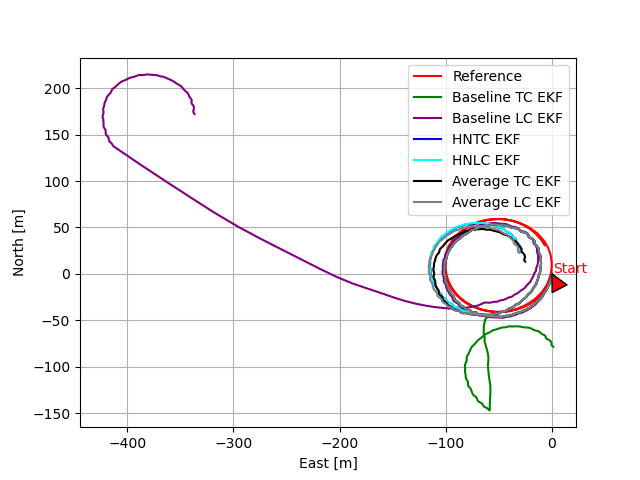}
    \caption{Trajectory 1 in the North-East plane.}
    \label{fig:sub1}
  \end{subfigure}
  \hfill
  \begin{subfigure}[b]{\columnwidth}
    \includegraphics[width=\columnwidth]{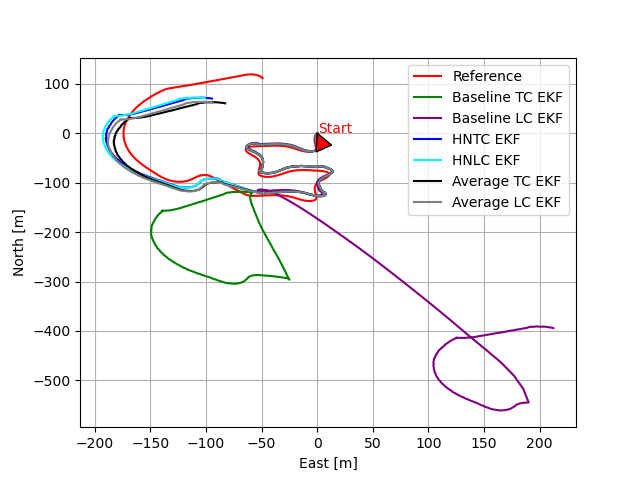}
    \caption{Trajectory 2 in the North-East plane.}
    \label{fig:sub2}
  \end{subfigure}
  \caption{Visualization of the trajectory of all inspected approaches with respect to the reference trajectory in red, plotted in the North-East plane. The model-based LC and TC approaches exhibit divergence during the 30-second period of missing beams, whereas the other methods maintain good results. The suggested approach outperforms both model-based methods.}
  \label{fig:all_pos}
\end{figure}

\noindent
Initially, we explored the different durations of partial DVL measurements, ranging from 5 seconds up to 25 seconds, on the filter performance. We observed that the performance of the model-based approaches was consistently satisfactory. This is attributed to the high quality of the inertial sensors. Hence, we extended our investigation to a 30-second window of partial beam measurements and evaluated the performance of both  LC and TC approaches in these scenarios. To ensure a fair comparison with the network's beam prediction, we applied also an average estimator to regress the missing beams. \hlcc{It is important to emphasize that the regressed beams are not actual DVL beam measurements. Thus, this should be reflected in different variance values in the measurement noise covariance matrix} $\mathbf{R}_{k}$. Moreover, as the network utilizes all $N$ \hlcc{past DVL beam measurements along with the current measured ones, it introduces correlations that should also be reflected in the measurement noise covariance matrix. However, for simplicity, we assume these effects are small and neglect them.}
In Fig. \ref{fig:velSleeve}, the behavior of the velocity error states in the EKF is presented. The error remains bounded within the estimated covariance sleeve (1 sigma), except for a few measurements that exceed it. Additionally, the estimated covariance by the EKF appears to be converging over time. This behavior is expected when all four beams are available, as indicated by the tight constraint on the velocity error state. However, in scenarios where the bottom lock condition is not met, leading to fewer available beams, the uncertainty in the velocity measurements increases, resulting in a wider covariance sleeve.
\begin{figure}[b!]
  \centering
  \begin{subfigure}[b]{\columnwidth}
    \includegraphics[width=\columnwidth]{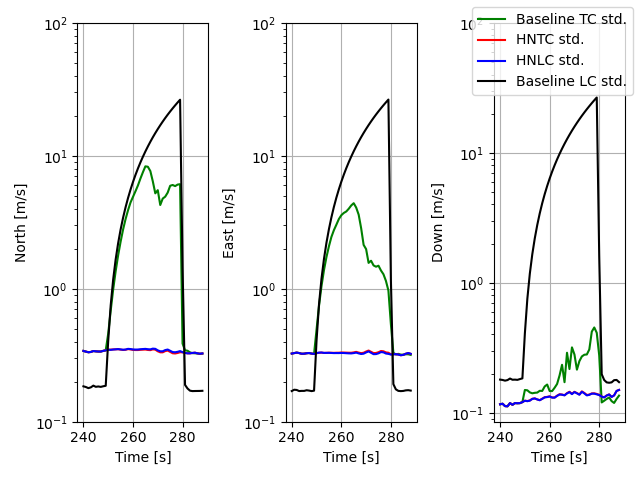}
    \caption{ The estimated velocity standard deviation for trajectory 1 over time.}
    \label{fig:sub1}
  \end{subfigure}
  \hfill
  \begin{subfigure}[b]{\columnwidth}
    \includegraphics[width=\columnwidth]{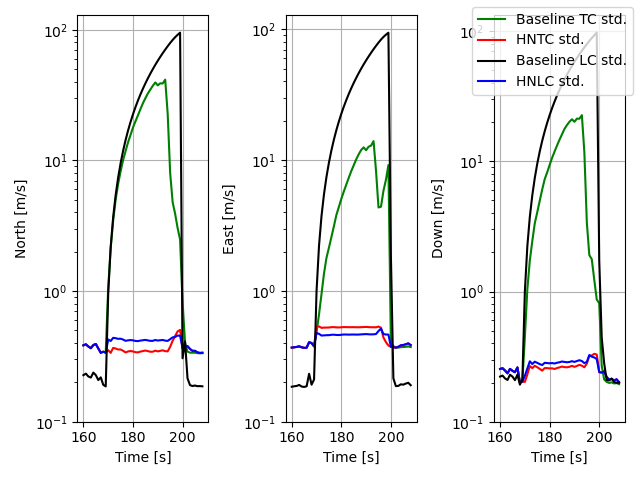}
    \caption{The estimated velocity standard deviation for trajectory 2 over time.}
    \label{fig:sub2}
  \end{subfigure}
  \caption{The estimated velocity standard deviation trajectory 1 (a) and trajectory 2 (b) over time, comparing the baseline LC and TC approaches to the suggested HNLC and HNTC approaches. The suggested approaches were able to maintain an almost constant standard deviation, whereas in the model-based approaches, it diverges. }
  \label{fig:cov}
\end{figure}
\\ \noindent
Figure \ref{fig:cov} displays a magnified perspective spanning a 30-second duration, depicting the covariance of velocity estimates derived from partial DVL measurements for instances where two beams are missing, covering both trajectory 1 and trajectory 2. This visualization elucidates the divergence of uncertainty during encounters with missing beam situations, followed by subsequent convergence once all the beams are available again. This behavior is also observed in the case of three missing beams, but as expected, with a higher divergence rate and final value. 
When two beams are missing, the baseline LC EKF operates solely on pure inertial navigation, as the velocity cannot be estimated using the LS parameter estimator. This explains the divergence of the estimated velocity covariance, represented by a black curve. On the other hand, in the baseline TC approach, partial DVL measurements are utilized to mitigate error accumulation. Although the partial measurements do not provide full velocity information, they offer more certainty compared to the LC approach, as indicated by the magnitude of the green curve. When examining the HNLC and HNTC approaches, no divergence is observed, and the standard deviation values in both the neural loosely and neural tightly coupled methods are similar, with only a small difference as illustrated in Fig.\ref{fig:cov}.
\\ \noindent
All of the factors mentioned above contribute to a significant improvement in VRMSE. Our HNLC and HNTC approaches outperform the model-based methods in both cases of two and three missing beams. Specifically, for two missing beams, the proposed approach demonstrates an improvement of over 93.6\%, and for three missing beams, more than 94.9\% in terms of VRTE. We also compared our approach to an average estimator, which takes the past DVL beam measurements and calculates the average to forecast the missing beam. Then, utilizes it in the LC or TC methods to update the EKF. This comparison is considered fair as both methods receive the same inputs. The average estimator yields good results primarily because the AUV's maneuvers are not extreme and do not change frequently. However, the neural coupled approach is able to outperform the estimator in terms of VRMSE. With two missing beams, the proposed approach demonstrates an improvement of over 5.8\% for two missing beams and over 11.4\% for three missing beams. All the results are summarized in Table \ref{tab:res}, and the performance of all the methods in terms of position accuracy in the North-East plane is illustrated in Fig.\ref{fig:all_pos}.

\begin{table*}[t!]
\centering
\caption{A comparison of the HNLC and HNTC approaches' performance with respect to the model-based LC and TC baseline, and the model-based beam regressor (average estimator) in terms of both the VRMSE and the VRTE.}
\label{tab:res}
\resizebox{\textwidth}{!}{%
\begin{tabular}{cccccccc}
\hline
\multirow{2}{*}{Scenario} &
  \multirow{2}{*}{Method} &
  \multicolumn{2}{c}{VRMSE {[}m/s{]}} &
  \multicolumn{2}{c}{VRTE w.r.t Baseline} &
  \multicolumn{2}{c}{\begin{tabular}[c]{@{}c@{}}VRTE w.r.t Average\\ Estimator\end{tabular}} \\
 &
   &
  Trajectory 1 &
  Trajectory 2 &
  Trajectory 1 &
  Trajectory 2 &
  Trajectory 1 &
  Trajectory 2 \\ \hline
\multirow{6}{*}{2 Missing Beams} & Baseline LC         & 8.62 & 12.01 & N/A     & N/A     & N/A     & N/A     \\
                                 & Baseline TC         & 2.80 & 6.08  & N/A     & N/A     & N/A     & N/A     \\
                                 & Average LC       & 0.19 & 0.42  & 97.7 \% & 96.5 \% & N/A     & N/A     \\
                                 & Average TC       & 0.18 & 0.41  & 93.2 \% & 93.1 \% & N/A     & N/A     \\
                                 & HNLC (Ours) & 0.17 & 0.35  & 97.9 \% & 97.0 \% & 10.6 \% & 14.5 \% \\
                                 & HNTC (Ours) & 0.17 & 0.35  & 93.6 \% & 94.1 \% & 5.8 \%  & 14.2 \% \\ \hline
\multirow{6}{*}{3 Missing Beams} & Baseline LC         & 9.53 & 11.23 & N/A     & N/A     & N/A     & N/A     \\
                                 & Baseline TC         & 5.17 & 10.53 & N/A     & N/A     & N/A     & N/A     \\
                                 & Average LC       & 0.28 & 0.35  & 96.9 \% & 96.8 \% & N/A     & N/A     \\
                                 & Average TC       & 0.29 & 0.34  & 94.3 \% & 96.6 \% & N/A     & N/A     \\
                                 & HNLC (Ours) & 0.25 & 0.29  & 97.3 \% & 97.3 \% & 11.4 \% & 15.8 \% \\
                                 & HNTC (Ours) & 0.25 & 0.29  & 94.9 \% & 97.1 \% & 12.2 \% & 14.8 \% \\ \hline
\end{tabular}%
}
\end{table*}

\section{Conclusions}\label{Con}
\noindent
The underwater environment poses significant challenges for navigation due to its harsh conditions and the unavailability of GNSS signals for precise position updates. Typically, navigation systems integrate an INS with a DVL using an EKF. The DVL relies on a condition known as bottom lock, which requires at least three received beams out of the four transmitted. When fewer beams are available, the navigation solution suffers, whether in a TC or LC approach. In this paper, we proposed the HNLC and HNTC approaches to cope with limited beam scenarios. We designed a deep learning architecture to regress the missing beams using past beam measurements. In addition to comparing the results to the baseline model-based filter performance, we also compared it to an average estimator, which receives the same inputs as our suggested network. Our approach was evaluated on real-world AUV data from two different occasions to emphasize its robustness. The results demonstrate that the suggested approach outperforms the model-based approaches by an average of 95.65\% and the performance of the average estimator by an average of 11.27\% in scenarios with two missing beams. In the case of three missing beams, it outperforms the model-based approaches by an average of 96.65\% and the performance of the average estimator by an average of 13.55\%. Our proposed approach requires only software modifications to enable seamless AUV navigation, even in situations of limited beam measurement.
\\ \noindent
In future work, we intend to delve deeper into investigating the cross-correlations that arise from using DL approaches with past and current DVL beam measurements, particularly in how they impact the measurements' noise covariance matrix. Additionally, prior research has indicated that incorporating inertial data alongside past and current DVL measurements may lead to improved beam and velocity estimation. However, this approach introduces additional correlations between the process data and the measurement data, which must be carefully addressed when integrating them into the EKF. We aim to analyze the effects of these correlations on performance in future studies.

\section*{Acknowledgment}
\noindent
N.C. is supported by the Maurice Hatter Foundation and
University of Haifa presidential scholarship for outstanding students on a direct Ph.D. track.

\bibliographystyle{IEEEtran}
\bibliography{bio.bib}

\end{document}